\documentclass{article}

\usepackage[final, nonatbib]{nips_2016}
\usepackage[usenames, dvipsnames]{color}

\usepackage[utf8]{inputenc} 
\usepackage[T1]{fontenc}    
\usepackage{hyperref}       
\usepackage{url}            
\usepackage{booktabs}       
\usepackage{amsfonts}       
\usepackage{nicefrac}       
\usepackage{microtype}      
\usepackage{bm}
\usepackage{amsmath}
\usepackage{graphicx}
\usepackage{tabularx,booktabs}

\title{Parallel WaveNet: \\Fast High-Fidelity Speech Synthesis}

\author{
Aaron van den Oord, Yazhe Li, Igor Babuschkin, Karen Simonyan, Oriol Vinyals,\\
\textbf{Koray Kavukcuoglu}\\
\texttt{avdnoord,yazhe,ibab,simonyan,vinyals,korayk@google.com}\\
\AND
George van den Driessche, Edward Lockhart, Luis C. Cobo, Florian Stimberg,\\
\textbf{Norman Casagrande, Dominik Grewe, Seb Noury, Sander Dieleman, Erich Elsen,}\\
\textbf{Nal Kalchbrenner, Heiga Zen, Alex Graves, Helen King, Tom Walters, Dan Belov,}\\
\textbf{Demis Hassabis}
}

\begin{document}

\maketitle

\begin{abstract}
The recently-developed \emph{WaveNet} architecture~\cite{wavenet2016} is the current state of the art in realistic speech synthesis, consistently rated as more natural sounding for many different languages than any previous system.
However, because WaveNet relies on sequential generation of one audio sample at a time, it is poorly suited to today's massively parallel computers, and therefore hard to deploy in a real-time production setting.
This paper introduces \emph{Probability Density Distillation}, a new method
for training a parallel feed-forward network from a trained WaveNet with no significant difference in quality.
The resulting system is capable of generating high-fidelity speech samples at more than 20 times faster than real-time, and is deployed online by Google Assistant, including serving multiple English and Japanese voices.
\end{abstract}

\section{Introduction}

Recent successes of deep learning go beyond achieving state-of-the-art results in research benchmarks, and push the frontiers in some of the most challenging real world applications such as speech recognition \cite{hinton2012deep}, image recognition \cite{krizhevsky2012imagenet,szegedy2015going}, and machine translation \cite{wu2016google}. 
The recently published WaveNet~\cite{wavenet2016} model achieves state-of-the-art results in speech synthesis, and significantly closes the gap with natural human speech. However, it is not well suited for real world deployment due to its prohibitive generation speed. In this paper, we present a new algorithm for distilling WaveNet into a feed-forward neural network which can synthesise equally high quality speech much more efficiently, and is deployed to millions of users.

WaveNet is one of a family of autoregressive deep generative models that have been applied with great success to data as diverse as text~\cite{mikolov2010recurrent}, images~\cite{larochelle2011neural,theis2015generative,oord2016pixel,van2016conditional}, video~\cite{kalchbrenner2016video}, handwriting~\cite{graves2013generating} as well as human speech and music.
Modelling raw audio signals, as WaveNet does, represents a particularly extreme form of autoregression, with up to 24,000 samples predicted per second.
Operating at such a high temporal resolution is not problematic during network training, where the complete sequence of input samples is already available and---thanks to the convolutional structure of the network---can be processed in parallel.
When generating samples, however, each input sample must be drawn from the output distribution before it can be passed in as input at the next time step, making parallel processing impossible.

Inverse autoregressive flows (IAFs)~\cite{kingma2016improving} represent a kind of dual formulation of deep autoregressive modelling, in which sampling can be  performed in parallel, while the inference procedure required for likelihood estimation is sequential and slow.
The goal of this paper is to marry the best features of both models: the efficient training of WaveNet and the efficient sampling of IAF networks.
The bridge between them is a new form of neural network distillation~\cite{hinton2015distilling}, which we refer to as \emph{Probability Density Distillation}, where a trained WaveNet model is used as a teacher for a feedforward IAF model.

The next section describes the original WaveNet model, while Sections 3 and 4 define in detail the new, parallel version of WaveNet and the distillation process used to transfer knowledge between them. 
Section 5 then presents experimental results showing no loss in perceived quality for parallel versus original WaveNet, and continued superiority over previous benchmarks. 
We also present timings for sample generation, demonstrating more than 1000$\times$ speed-up relative to original WaveNet.

\section{WaveNet}
\label{sec:wavenet}

Autoregressive networks model the joint distribution of high-dimensional data as a product of conditional distributions using the probabilistic chain-rule:
$$
p(\bm{x}) = \prod_t p(x_t|x_{<t}, \bm{\theta}),
$$
where $x_t$ is the $t$-th variable of $\bm{x}$ and $\bm{\theta}$ are the parameters of the autoregressive model. The conditional distributions are usually modelled with a neural network that receives $x_{<t}$ as input and outputs a distribution over possible $x_t$.

WaveNet \cite{wavenet2016} is a convolutional autoregressive model which produces all $p(x_t|x_{<t})$ in one forward pass, by making use of \emph{causal}---or \emph{masked}---convolutions \cite{oord2016pixel, germain2015made}. 
Every causal convolutional layer can process its input in parallel, making these architectures very fast to train compared to RNNs~\cite{van2016conditional}, which can only be updated sequentially. 
At generation time, however, the waveform has to be synthesised in a sequential fashion as $x_t$ must be sampled first in order to obtain $x_{>t}$.
Due to this nature, real time (or faster) synthesis with a fully autoregressive system is challenging.
While sampling speed is not a significant issue for offline generation, it is essential for real-word applications.
A version of WaveNet that generates in real-time has been developed~\cite{paineFastWaveNet}, but it required the use of a much smaller network, resulting in severely degraded quality.

\begin{figure}[ht]
\centering
\includegraphics[trim={6.625in 0 4.95in 0},clip,width=\linewidth]{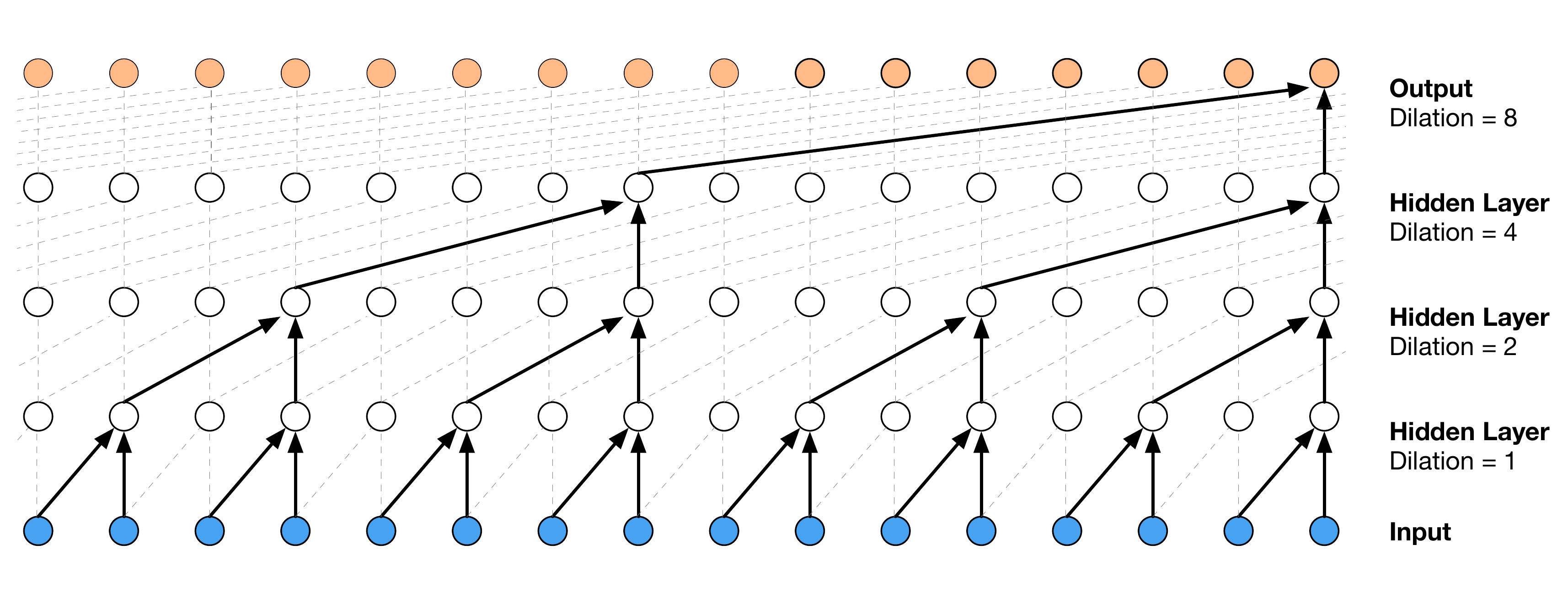}
\caption{Visualisation of a WaveNet stack and its receptive field \cite{wavenet2016}.}
\label{fig:masked_dilated_convolution}
\end{figure}

Raw audio data is typically very high-dimensional (e.g. 16,000 samples per second for 16kHz audio), and contains complex, hierarchical structures spanning many thousands of time steps, such as words in speech or melodies in music.
Modelling such long-term dependencies with standard causal convolution layers would require a very deep network to ensure a sufficiently broad receptive field. 
WaveNet avoids this constraint by using \emph{dilated} causal convolutions, which allow the receptive field to grow exponentially with depth.

WaveNet uses gated activation functions, together with a simple mechanism introduced in \cite{oord2016pixel} to condition on extra information such as class labels or linguistic features:
\begin{equation}
\bm{h}_i = \sigma\left(W_{g,i} \ast \bm{x}_i + V_{g,i}^T \bm{c}\right) \odot \tanh\left(W_{f,i} \ast \bm{x}_i + V_{f,i}^T \bm{c}\right),
\label{activation}
\end{equation}
where $\ast$ denotes a convolution operator, and $\odot$ denotes an element-wise multiplication operator. $\sigma(\cdot)$ is a logistic sigmoid function. $\bm{c}$ represents extra conditioning data. $i$ is the layer index. $f$ and $g$ denote filter and gate, respectively. $W$ and $V$ are learnable weights. 
In cases where $\bm{c}$ encodes spatial or sequential information (such as a sequence of linguistic features), the matrix products ($V_{f,i}^T \bm{c}$ and $V_{g,i}^T \bm{c}$) are replaced by convolutions ($V_{f,i} \ast \bm{c}$ and $V_{g,i} \ast \bm{c}$).

\subsection{Higher Fidelity WaveNet}
\label{sec:high_fidelity}
For this work we made two improvements to the basic WaveNet model to enhance its audio quality for production use.
Unlike previous versions of WaveNet \cite{wavenet2016}, where 8-bit ($\mu$-law or PCM) audio was modelled with a 256-way categorical distribution, we increased the fidelity by modelling 16-bit audio. 
Since training a 65,536-way categorical distribution would be prohibitively costly, we instead modelled the samples with the discretized mixture of logistics distribution introduced in \cite{salimans2017pixelcnn}.
We further improved fidelity by increasing the audio sampling rate from 16kHz to 24kHz.
This required a WaveNet with a wider receptive field, which we achieved by increasing the dilated convolution filter size from 2 to 3.
An alternative strategy would be to increase the number of layers or add more dilation stages.

\section{Parallel WaveNet}
\label{sec:arch}
While the convolutional structure of WaveNet allows for rapid parallel training, sample generation remains inherently sequential and therefore slow, as it is for all autoregressive models which use ancestral sampling.
We therefore seek an alternative architecture that will allow for rapid, parallel generation.

Inverse-autoregressive flows (IAFs) \cite{kingma2016improving} are stochastic generative models whose latent variables are arranged so that all elements of a high dimensional observable sample can be generated in parallel.
IAFs are a special type of normalising flow \cite{dinh2014nice, rezende2015variational, dinh2016density} which model a multivariate distribution $p_X(\bm{x})$ as an explicit invertible non-linear transformation $f$ of a simple tractable distribution $p_Z(\bm{z})$ (such as an isotropic Gaussian distribution). The resulting random variable $\bm{x} = f(\bm{z})$ has a log probability:
$$
\log p_X(\bm{x}) = \log p_Z(\bm{z}) - \log \Big| \frac{d\bm{x}}{d\bm{z}} \Big|,
$$
where $\big| \frac{d\bm{x}}{d\bm{z}} \big|$ is the determinant of the Jacobian of $f$. 
The transformation $f$ is typically chosen so that it is invertible and its Jacobian determinant is easy to compute.
In the case of an IAF, $x_t$ is modelled by $p(x_t| \bm{z}_{\leq t})$ so that $x_t = f(\bm{z}_{\leq t})$. The transformation has a triangular Jacobian matrix which makes the determinant simply the product of the diagonal entries:
$$
\log \Big| \frac{d\bm{x}}{d\bm{z}} \Big| = \sum_t \log \frac{\partial f(\bm{z}_{\leq t})}{\partial z_t}.
$$
Initially, a random sample is drawn from $\bm{z} \sim \mathrm{Logistic}(0, I)$. The following transformation is applied to $\bm{z}$:
\begin{align}
x_t = z_t \cdot s(\bm{z}_{<t}, \bm{\theta}) + \mu(\bm{z}_{<t}, \bm{\theta})
\end{align}
The network outputs a sample $\bm{x}$, as well as $\bm{\mu}$ and $\bm{s}$. Therefore, $p(x_t|\bm{z}_{<t})$ follows a logistic distribution parameterised by $\mu_t$ and $s_t$.
$$
p(x_t|\bm{z}_{<t}, \bm{\theta})=\mathop{\mathbb{L}}\left(x_t \big| \mu(\bm{z}_{<t}, \bm{\theta}), s(\bm{z}_{<t}, \bm{\theta}) \right),
$$ 
While $\mu(\bm{z}_{<t}, \bm{\theta})$ and $s(\bm{z}_{<t}, \bm{\theta})$ can be any autoregressive model, we use the same convolutional autoregressive network structure as the original WaveNet~\cite{wavenet2016}. 
If an IAF and an autoregressive model share the same output distribution class (e.g., mixture of logistics or categorical) then mathematically they should be able to model the same multivariate distributions. However, in practice there are some differences (see Appendix section \ref{sec:remarks_IAF}).
To output the correct distribution for timestep $x_t$, the inverse autoregressive flow can implicitly infer what it would have output at previous timesteps $x_1, \dots, x_{t-1}$ based on the noise inputs $z_1, \dots, z_{t-1}$, which allows it to output all $x_t$ in parallel given $z_t$. 

In general, normalising flows might require repeated iterations to transform uncorrelated noise into structured samples, with the output generated by the flow at each iteration passed in as input at the next~\cite{rezende2015variational} one.
This is less crucial for IAFs, as the autoregressive latents can induce significant structure in a single pass.
Nonetheless we observed that having up to 4 flow iterations (which we implemented by simply stacking 4 such networks on top of each other) did improve the quality.
Note that in the final parallel WaveNet architecture, the weights were not shared between the flows.

The first (bottom) network takes as input the white unconditional logistic noise: $\bm{x}^0 = \bm{z}$.
Thereafter the output of each network $i$ is passed as input to the next network $i+1$ , which again transforms it.
\begin{align}
\bm{x}^i &=\bm{x^{i-1}} \cdot \bm{s}^i + \bm{\mu}^i
\end{align}
Because we use the same ordering in all the flows, the final distribution $p(x_t|z_{<t}, \bm{\theta})$ is logistic with location $\mu_\text{tot}$ and scale $s_\text{tot}$:
\begin{align}
\bm{\mu}_\text{tot} &= \sum_i^N \bm{\mu}^i \left(\prod_{j>i}^N \bm{s}^j\right)\\
\bm{s}_\text{tot}  &= \prod_i^N \bm{s}_i
\end{align}
where $N$ is the number of flows and the dependencies on $t$ and $z$ are omitted for simplicity.

\section{Probability Density Distillation}
\label{sec:kl_loss}

Training the parallel WaveNet model directly with maximum likelihood would be impractical, as the inference procedure required to estimate the log-likelihoods is sequential and slow\footnote{In this sense the two architectures are dual to one another: slow training and fast generation with parallel WaveNet versus fast training and slow generation with WaveNet.}.
We therefore introduce a novel form of neural network distillation~\cite{hinton2015distilling} that uses an already trained WaveNet as a `teacher' from which a parallel WaveNet `student' can efficiently learn.
To stress the fact that we are dealing with normalised density models, we refer to this process as \emph{Probability Density Distillation} (in contrast to Probability Density Estimation).
The basic idea is for the student to attempt to match the probability of its own samples under the distribution learned by the teacher.

Given a parallel WaveNet student $p_S(\bm{x})$ and WaveNet teacher $p_T(\bm{x})$ which has been trained on a dataset of audio, we define the \emph{Probability Density Distillation} loss as follows:
\begin{equation}
D_{\text{KL}}\left(P_S|| P_T\right)=H(P_S, P_T) - H(P_S) \label{KLD}
\end{equation}
where $D_{\text{KL}}$ is the Kullback–Leibler divergence, and $H(P_S, P_T)$ is the cross-entropy between the student $P_S$ and teacher $P_T$, and $H(P_S)$ is the entropy of the student distribution.
When the KL divergence becomes zero, the student distribution has fully recovered the teacher's distribution.
The entropy term (which is not present in previous distillation objectives~\cite{hinton2015distilling}) is vital in that it prevents the student's distribution from collapsing to the mode of the teacher (which, counter-intuitively, does not yield a good sample---see Appendix section \ref{sec:remarks_MAP}).
Crucially, all the operations required to estimate derivatives for this loss (sampling from $p_S(\bm{x})$, evaluating $p_T(\bm{x})$, and evaluating $H(P_S)$) can be performed efficiently, as we will see.

It is worth noting the parallels to Generative Adversarial Networks (GANs~\cite{goodfellow2014generative}), with the student playing the role of generator, and the teacher playing the role of discriminator.
As opposed to GANs, however, the student is not attempting to fool the teacher in an adversarial manner; rather it co-operates by attempting to match the teacher's probabilities. 
Furthermore the teacher is held constant, rather than being trained in tandem with the student, and both models yield tractable normalised distributions.

Recently~\cite{gu2017natmt} has presented a related idea to train feed-forward networks for neural machine translation. Their method is based on conditioning the feedforward decoder on fertility values, which require supervision by an external alignment system. The training procedure also involves the creation of an additional dataset as well as fine-tuning. During inference, their model relies on re-scoring by an auto-regressive model.

\begin{figure}[ht]
\centering
\includegraphics[width=\linewidth]{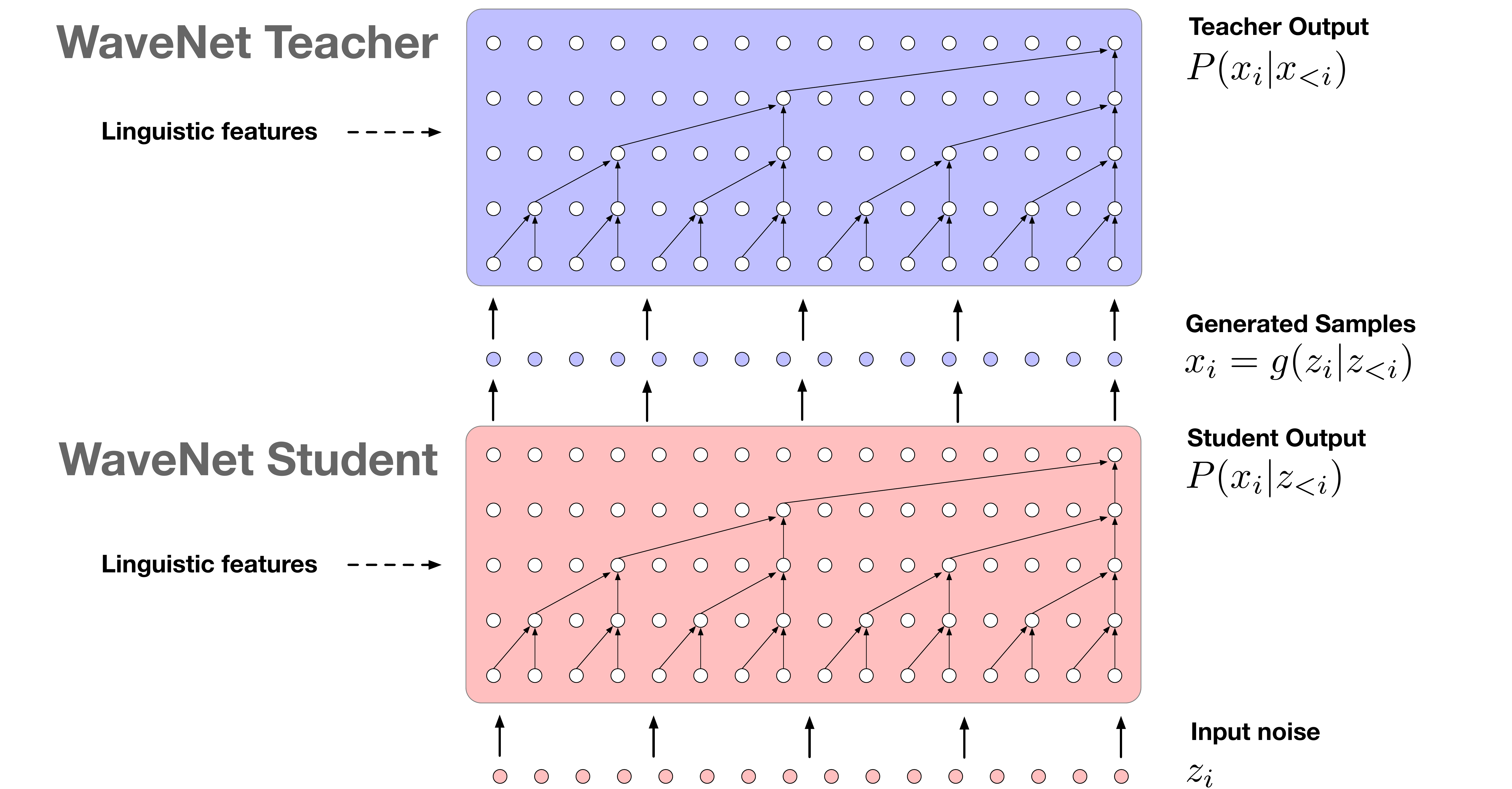}
\caption{\textbf{Overview of Probability Density Distillation}. A pre-trained  WaveNet teacher is used to score the samples $\bm{x}$ output by the student. The student is trained to minimise the KL-divergence between its distribution and that of the teacher by maximising the log-likelihood of its samples under the teacher and maximising its own entropy at the same time.}
\label{fig:distillation}
\end{figure}

First, observe that the entropy term $H(P_S)$ in Equation \ref{KLD} can be rewritten as follows:
\begin{align}
H(P_S) &= \mathop{\mathbb{E}}_{z\sim L(0, 1)}\left[\sum_{t=1}^T - \ln p_S(x_t|\bm{z}_{<t})\right] \label{entropy1}\\
&= \mathop{\mathbb{E}}_{z\sim L(0, 1)}\left[\sum_{t=1}^T \ln s(\bm{z}_{<t}, \bm{\theta})\right] + 2T,  \label{entropy2}
\end{align}
where $\bm{x}=g(\bm{z})$ and $z_t$ are independent samples drawn from the logistic distribution. The second equality in Equation \ref{entropy2} follows because the entropy of a logistic distribution $\mathop{\mathbb{L}}(\mu, s)$ is $\ln s + 2$.
We can therefore compute this term without having to explicitly generate $\bm{x}$.

The cross-entropy term $H(P_S, P_T)$ however explicitly depends on $\bm{x}=g(\bm{z})$, and therefore requires sampling from the student to estimate.
\begin{align}
H(P_S, P_T) 
&= \int_{\bm{x}} p_S(\bm{x}) \ln p_T(\bm{x})  \label{eq:vanilla_ce} \\
&= \sum_{t=1}^T \int_{\bm{x}}p_S(\bm{x}) \ln p_T(x_t|\bm{x}_{<t}) \\
&= \sum_{t=1}^T \int_{\bm{x}}p_S(\bm{x}_{<t})p_S(x_t|\bm{x}_{<t})p_S(\bm{x}_{>t}|\bm{x}_{\leq t}) \ln p_T(x_t|\bm{x}_{<t}) \\
&= \sum_{t=1}^T \mathop{\mathbb{E}}_{p_S(\bm{x}_{<t})} \bigg[\int_{x_t} p_S(x_t|\bm{x}_{<t}) \ln p_T(x_t|\bm{x}_{<t})\int_{\bm{x}_{>t}} p_S(\bm{x}_{>t}|\bm{x}_{\leq t})\bigg] \\
&= \sum_{t=1}^T \mathop{\mathbb{E}}_{p_S(\bm{x}_{<t})} H\Big(p_S(x_t|\bm{x}_{<t}), p_T(x_t|\bm{x}_{<t})\Big).
\end{align}
For every sample $\bm{x}$ we draw from the student $p_S$ we can compute all $p_T(x_t|\bm{x}_{<t})$ in parallel with the teacher and then evaluate $H(p_S(x_t|\bm{x}_{<t}), p_T(x_t|\bm{x}_{<t}))$ very efficiently by drawing multiple different samples $x_t$ from $p_S(x_t|\bm{x}_{<t})$ for each timestep.
This unbiased estimator has a much lower variance than naively evaluating the sample under the teacher with Equation \ref{eq:vanilla_ce}.

Because the teacher's output distribution $p_T(x_t|\bm{x}_{<t})$ is parameterised as a mixture of logistics distribution, the loss term $\ln p_T(x_t|\bm{x}_{<t})$ is differentiable with respect to both $x_t$ and $x_{<t}$.
A categorical distribution, on the other hand, would only be differentiable w.r.t. $\bm{x}_{<t}$.

\subsection{Additional loss terms}
\label{sec:additional_loss}
Training with \emph{Probability Density Distillation} alone might not sufficiently constrain the student to generate high quality audio streams. Therefore, we also introduce additional loss functions to guide the student distribution towards the desired output space.

\subsubsection*{Power loss}
\label{sec:power_loss}

The first additional loss we propose is the \emph{power loss}, which ensures that the power in different frequency bands of the speech are on average used as much as in human speech. The power loss helps to avoid the student from collapsing to a high-entropy WaveNet-mode, such as whispering.

The power-loss is defined as:
\begin{equation}
||\phi(g(\bm{z}, \bm{c})) - \phi(\bm{y})||^2, \label{powerloss}
\end{equation}
where $(\bm{y}, \bm{c})$ is an example with conditioning from the training set, $\phi(\bm{x})=|\text{STFT}(\bm{x})|^2$ and STFT stands for the Short-Term Fourier Transform. We found that $\phi(\bm{x})$ can be averaged over time before taking the Euclidean distance with little difference in effect, which means it is the \emph{average} power for various frequencies that is important.

\subsubsection*{Perceptual loss}
\label{sec:perceptual_loss}
In the power loss formulation given in equation~\ref{powerloss}, one can also use a neural network instead of the STFT to conserve a perceptual property of the signal rather than total energy. In our case we have used a WaveNet-like classifier trained to predict the phones from raw audio. Because such a classifier naturally extracts high-level features that are relevant for recognising the phones, this loss term penalises bad pronunciations. A similar principle has been used in computer vision for artistic style transfer \cite{gatys2015neural}, or to get better perceptual reconstruction losses, e.g., in super-resolution \cite{johnson2016perceptual}.

We have experimented with two different ways of using the perceptual loss, the feature reconstruction loss (the Euclidean distance between feature maps in the classifier) and the style loss (the Euclidean distance between the Gram matrices~\cite{johnson2016perceptual}). The latter produced better results in our experiments. 

\subsubsection*{Contrastive loss}
\label{sec:contrastive_loss}

Finally, we also introduce a contrastive distillation loss as follows:

\begin{equation}
D_\text{KL}\Big(P_S(\bm{c}_1)\Big|\Big|P_T(\bm{c}_1)\Big)-\gamma D_\text{KL}\Big(P_S(\bm{c}_1)\Big|\Big|P_T\bm{c}_2)\Big),
\end{equation}

which minimises the KL-divergence between the teacher and student when both are conditioned on the same information $\bm{c}_1$ (e.g., linguistic features, speaker ID, \dots), but also maximises it for different conditioning pairs $\bm{c}1\neq \bm{c}2$. In order to implement this loss, we use the output of the student $\bm{x}=g(\bm{z}, \bm{c}_1)$ and evaluate the waveform twice under the teacher: once with the same conditioning $P_T(\bm{x}|\bm{c}_1)$ and once with a randomly sampled conditioning input: $P_T(\bm{x}|\bm{c}_2)$. The weight for the contrastive term $\gamma$ was set to 0.3 in our experiments. The contrastive loss penalises waveforms that have high likelihood  regardless of the conditioning vector.

\section{Experiments}

In all our experiments we used text-to-speech models that were conditioned on linguistic features (similar to \cite{wavenet2016}), providing phonetic and duration information to the network. We also conditioned the models on pitch information (logarithm of $f_0$, the fundamental frequency) predicted by a different model. We never used ground-truth information (such as pitch or duration) extracted from human speech for generating audio samples and the test sentences were not present (or similar to those) in the training set.

The teacher WaveNet network was trained for 1,000,000 steps with the ADAM optimiser \cite{kingma2014adam} with a minibatch size of 32 audio clips, each containing 7,680 timesteps (roughly 320ms). Remarkably, a relatively short snippet of time is sufficient to train the parallel WaveNet to produce long term coherent waveforms. The learning rate was held constant at $2\times 10^{-4}$, and Polyak averaging \cite{polyak1992acceleration} was applied over the parameters. The model consists of 30 layers, grouped into 3 dilated residual block stacks of 10 layers. In every stack, the dilation rate increases by a factor of 2 in every layer, starting with rate 1 (no dilation) and reaching the maximum dilation of 512 in the last layer. The filter size of causal dilated convolutions is 3. The number of hidden units in the gating layers is 512 (split into two groups of 256 for the two parts of the activation function~\eqref{activation}). The number of hidden units in the residual connection is 512, and in the skip connection and the $1\times 1$ convolutions before the output layer is also 256. We used 10 mixture components for the mixture of logistics output distribution.

The student network consisted of the same WaveNet architecture layout, except with different inputs and outputs and no skip connections. The student was also trained for 1,000,000 steps with the same optimisation settings. The student typically consisted of 4 flows with 10, 10, 10, 30 layers respectively, with 64 hidden units for the residual and gating layers.

\subsubsection*{Audio Generation Speed}

We have benchmarked the sampling speed of autoregressive and distilled WaveNets on an NVIDIA P100 GPU. Both models were implemented in Tensorflow \cite{abadi2016tensorflow} and compiled with XLA. The hidden layer activations  from previous timesteps in the autoregressive model were cached with circular buffers \cite{paineFastWaveNet}. The resulting sampling speed with this implementation is \textbf{172 timesteps/second}  for a minibatch of size 1. The distilled model, which is more parallelizable, achieves \textbf{over 500,000 timesteps/second} with same batch size of 1, resulting in three orders of magnitude speed-up.

\begin{table}[t]
  \centering
  \begin{tabularx}{0.7\textwidth}{l|c}
    \toprule
    \textbf{Method} & \textbf{Subjective 5-scale MOS} \\
    \midrule\midrule
    \textbf{16kHz, 8-bit $\bm{\mu}$-law, 25h data}: \\
    LSTM-RNN parametric \cite{wavenet2016} & 3.67 $\pm$ 0.098 \\ 
    HMM-driven concatenative \cite{wavenet2016} & 3.86 $\pm$ 0.137 \\
    WaveNet \cite{wavenet2016} & 4.21 $\pm$ 0.081 \\
    \midrule
    \textbf{24kHz, 16-bit linear PCM, 65h data}: \\
    HMM-driven concatenative & 4.19 $\pm$ 0.097 \\
    Autoregressive WaveNet & 4.41 $\pm$ 0.069 \\
    Distilled WaveNet &  4.41 $\pm$ 0.078 \\
    \bottomrule
  \end{tabularx}
  \vspace{5pt}
  \caption{Comparison of WaveNet distillation with the autoregressive teacher WaveNet, unit-selection (concatenative), and previous results from \cite{wavenet2016}. MOS stands for Mean Opinion Score.}
  \label{tab:overview}
\end{table}%

\subsubsection*{Audio Fidelity}

In our first set of experiments, we looked at the quality of WaveNet distillation compared to the autoregressive WaveNet teacher and other baselines on data from a professional female speaker \cite{wavenet2016}. Table \ref{tab:overview} gives a comparison of autoregressive WaveNet, distilled WaveNet and current production systems in terms of mean opinion score (MOS). There is no difference between MOS scores of the distilled WaveNet ($4.41 \pm 0.08$) and autoregressive WaveNet ($4.41 \pm 0.07$), and both are significantly better than the concatenative unit-selection baseline ($4.19 \pm 0.1$).

It is also important to note that the difference in MOS scores of our WaveNet baseline result $4.41$ compared to the previous reported result $4.21$ \cite{wavenet2016} is due to the improvement in audio fidelity as explained in Section \ref{sec:high_fidelity}: modelling a sample rate of $24$kHz instead of $16$kHz and bit-depth of $16$-bit PCM instead of $8$-bit $\mu$-law.

\begin{table}[t]
  \centering
  \renewcommand{\arraystretch}{1.1}%
  \begin{tabularx}{1.0\textwidth}{l|c|c|c}
    \toprule
    & \textbf{Parametric} & \textbf{Concatenative} & \textbf{Distilled WaveNet} \\
    \midrule\midrule
    \textbf{English speaker 1} (female - 65h data) \quad  & 3.88 & 4.19 & 4.41 \\
    \textbf{English speaker 2} (male - 21h data)   & 3.96 & 4.09 &  4.34 \\
    \textbf{English speaker 3} (male - 10h data) & 3.77 & 3.65 &  4.47 \\
    \textbf{English speaker 4} (female - 9h data)  & 3.42 & 3.40 & 3.97 \\
    \textbf{Japanese speaker} (female - 28h data)  & 4.07 & 3.47 &  4.23 \\
    \bottomrule
  \end{tabularx}
  \vspace{5pt}
  \caption{Comparison of MOS scores on English and Japanese with multi-speaker distilled WaveNets.  Note that some speakers sounded less appealing to people and always get lower MOS, however distilled parallel WaveNet always achieved significantly better results.}
  \label{tab:speakers}
\end{table}%

\subsubsection*{Multi-speaker Generation}

By conditioning on the speaker-ids we can construct a single parallel WaveNet model that is able to generate multiple speakers' voices and their accents.
These networks require slightly more capacity than single speaker models and thus had 30 layers in each flow.
In Table \ref{tab:speakers} we show a comparison of such a distilled parallel WaveNet model with two main baselines: a parametric and a concatenative system.
In the comparison, we use a number of English speakers from a single model (one of them, English speaker 1, is the same speaker as in Table~\ref{tab:overview}) and a Japanese speaker from another model. For some speakers, the concatenative system gets better results than the parametric system, while for other speakers it is the opposite.
The parallel WaveNet model, on the other hand, significantly outperforms both baselines for all the speakers.

\subsubsection*{Ablation Studies}

\begin{table}[t]
  \centering
  \begin{tabularx}{0.95\textwidth}{l|c}
    \toprule
    & \textbf{Preference Scores} \\
    & \textbf{versus baseline concatenative system}\\
    \textbf{Method}& \textbf{Win - Lose - Neutral}\\
    \midrule\midrule
    \textbf{Losses used}                              & \\
    KL + Power                                        & 60\% - 15\% - 25\% \\
    KL + Power + Perceptual                           & 66\% - 10\% - 24\% \\
    KL + Power + Perceptual + Contrastive (= default) & 65\% - 9\%  - 26\% \\
    \bottomrule
  \end{tabularx}
  \vspace{5pt}
  \caption{Performance with respect to different combinations of loss terms. We report preference comparison scores since their mean opinion scores tend to be very close and inconclusive.}
  \label{tab:ablation_losses}
\end{table}

To analyse the importance of the loss functions introduced in Section \ref{sec:additional_loss} we show how the quality of the distilled WaveNet changes with different loss functions in Table \ref{tab:ablation_losses} (top). We found that MOS scores of these models tend to be very similar to each other (and similar to the result in Table \ref{tab:overview}). Therefore, we report subjective preference scores from a paired comparison test (“A/B test”), which we found to be more reliable for noticing small (sometimes qualitative) differences.  In these tests, the subjects were asked to listen to a pair of samples and choose which they preferred, though they could choose “neutral” if they did not have any preference.

As mentioned before, the KL loss alone does not constrain the distillation process enough to obtain natural sounding speech (e.g., low-volume audio suffices for the KL), therefore we do not report preference scores with only this term. The KL loss (section ~\ref{sec:kl_loss}) combined with power-loss is enough to generate quite natural speech. Adding the perceptual loss gives a small but noticeable improvement. Adding the contrastive loss does not improve the preference scores any further, but makes the generated speech less noisy, which is something most raters do not pay attention to, but is important for production quality speech.

As explained in Section \ref{sec:arch}, we use multiple inverse-autoregressive flows in the parallel WaveNet architecture: A model with a single flow gets a MOS score of 4.21, compared to a MOS score of 4.41 for models with multiple flows.

\section{Conclusion}
In this paper we have introduced a novel method for high-fidelity speech synthesis based on WaveNet~\cite{wavenet2016} using \emph{Probability Density Distillation}. The proposed model achieved several orders of magnitude speed-up compared to the original WaveNet with no significant difference in quality. Moreover, we have successfully transferred this algorithm to new languages and multiple speakers.

The resulting system has been deployed in production at Google, and is currently being used to serve Google Assistant queries in real time to millions of users\footnote{\url{https://deepmind.com/blog/wavenet-launches-google-assistant/}}. We believe that the same method presented here can be used in many different domains to achieve similar speed improvements whilst maintaining output accuracy.

\section{Acknowledgements}
In this paper, we have described the research advances that made it possible for WaveNet to meet the speed and quality requirements for being used in production at Google. At the same time, an equivalently significant effort has gone into integrating this new end-to-end deep learning model into the production pipeline, satisfying requirements of not only speed, but latency and reliability, among others. We would like to thank Ben Coppin, Edgar Duéñez-Guzmán, Akihiro Matsukawa, Lizhao Liu, Mahalia Miller, Trevor Strohman and Eddie Kessler for very useful discussions. We also would like to thank the entire DeepMind Applied and Google Speech teams for their foundational contributions to the project and developing the production pipeline.

\bibliography{main}
\bibliographystyle{plain}

\newpage
\appendix
\section{Appendix}

\subsection{Argument against MAP estimation}
\label{sec:remarks_MAP}

In this section we make an argument against \emph{maximum a posteriori} (MAP) estimation for distillation; similar arguments have been made by previous authors in a different setting \cite{sonderby2016amortised}.

The distillation loss defined in Section~\ref{sec:kl_loss} minimises the KL divergence between the teacher and generator.
We could instead have minimised only the cross-entropy between the teacher and generator (the standard distillation loss term~\cite{hinton2015distilling}), so that the samples by the generator are as likely as possible according to the teacher.
Doing so would give rise to MAP estimation. 
Counter-intuitively, audio samples obtained through MAP estimation do not sound as good as typical examples from the teacher: in fact they are almost completely silent, even if using conditional information such as linguistic features. 
This effect is not due to adversarial behaviour on the part of the teacher, but rather is a fundamental property of the data distribution which the teacher has approximated.

As an example consider the simple case where we have audio from a white random noise source: the distribution at every timestep is $\mathcal{N}(0, 1)$, regardless of the samples at previous timesteps. 
White noise has a very specific and perceptually recognizable sound: a continual hiss. 
The MAP estimate of this data distribution, and thus of any generative model that matches it well, recovers the distribution mode, which is 0 at every timestep: i.e. complete silence.
More generally, any highly stochastic process is liable to have a `noiseless' and therefore atypical mode.
For the KL divergence the optimum is to recover the full teacher distribution.
This is clearly different from any random sample from the distribution. 
Furthermore, if one changes the representation of the data (e.g., by nonlinearly pre-processing the audio signal), then the MAP estimate changes, unlike the KL-divergence in Equation \ref{KLD}, which is invariant to the coordinate system.

\subsection{Autoregressive Models and Inverse-autoregressive Flows}
\label{sec:remarks_IAF}

Although inverse-autoregressive flows (IAFs) and autoregressive models can in principle model the same distributions \cite{chen2016variational}, they have different inductive biases and may vary greatly in their capacity to model certain processes.
As a simple example consider the Fibonacci series $(1, 1, 2, 3, 5, 8, 13, \dots)$. 
For an autoregressive model this is easy to model with a receptive field of two: $f(k)=f(k-1)+f(k-2)$. 
For an IAF, however, the receptive field needs to be at least size $k$ to correctly model $k$ terms, leading to a larger model that is less able to generalise.

\end{document}